\documentclass[journal]{IEEEtran}

\ifCLASSINFOpdf
\else
\fi
\usepackage{amsmath}
\usepackage{url}
\usepackage{orcidlink}

\usepackage{multirow}
\usepackage{colortbl}
\usepackage{xcolor}
\usepackage{amsfonts}
\usepackage{graphicx}
\usepackage{pifont}
\usepackage{hyperref}
\usepackage{amssymb}

\begin{document}
\title{Kangaroo: A Powerful Video-Language Model Supporting Long-context Video Input}
%
%
%

\author{Jiajun~Liu$^{*}$, 
        Yibing~Wang$^{*}$,
        Hanghang~Ma$^{\dagger}$,
        Xiaoping~Wu$^{\dagger}$,
        Xiaoqi~Ma$^{\dagger}$,
        Xiaoming~Wei,
        Jianbin~Jiao,
        Enhua~Wu,
        and~Jie~Hu$^{\ddagger}$ 
\thanks{Jiajun~Liu, Hanghang~Ma, Xiaoping~Wu, Xiaoqi~Ma, and Xiaoming~Wei are with Meituan Group, Beijing 100102, China (E-mail: \{liujiajun18, mahanghang,  wuxiaoping03, maxiaoqi05, weixiaoming\}@meituan.com).}
\thanks{Yibing~Wang and Jianbin~Jiao are with the School of Electronic, Electrical and Communication Engineering, the University of Chinese Academy of Sciences, (UCAS), Beijing, 100049, China. (E-mail:~wangyibing18@mails.ucas.ac.cn).}
\thanks{Jie~Hu and Enhua~Wu are with the Key Laboratory of System Software (Chinese Academy of Sciences) and State Key Laboratory of Computer Science, Institute of Software, Chinese Academy of Sciences, Beijing, 100190, China, and also with the University of Chinese Academy of Sciences, Beijing, 100049, China.
Jie Hu is also with Meituan Group (E-mail:~hujie@ios.ac.cn; ehwu@um.edu.mo).}
\thanks{The code and model are openly accessible to the public and can be found at \textit{https://github.com/KangarooGroup/Kangaroo}.}
\thanks{* Joint first authors. $\dagger$ Key contributors.}
\thanks{$\ddagger$ Corresponding author and Project lead.}
}

\maketitle

\begin{abstract}
Rapid advancements have been made in extending Large Language Models (LLMs) to Large Multi-modal Models (LMMs). However, extending input modality of LLMs to video data remains a challenging endeavor, especially for long videos. Due to insufficient access to large-scale high-quality video data and the excessive compression of visual features, current methods exhibit limitations in effectively processing long videos. In this paper, we introduce \textbf{Kangaroo}, a powerful Video LMM aimed at addressing these challenges. Confronted with issue of inadequate training data, we develop a data curation system to build a large-scale dataset with high-quality annotations for vision-language pre-training and instruction tuning. In addition, we design a curriculum training pipeline with gradually increasing resolution and number of input frames to accommodate long videos. Evaluation results demonstrate that, with 8B parameters, Kangaroo achieves state-of-the-art performance across a variety of video understanding benchmarks while exhibiting competitive results on others. Particularly, on benchmarks specialized for long videos, Kangaroo excels some larger models with over 10B parameters and proprietary models.
\end{abstract}



%
\IEEEpeerreviewmaketitle

\section{Introduction}
\begin{figure}[htbp]
  \centering
  \includegraphics[trim=0 6.4cm 13.2cm 0, clip, width=0.99\linewidth]{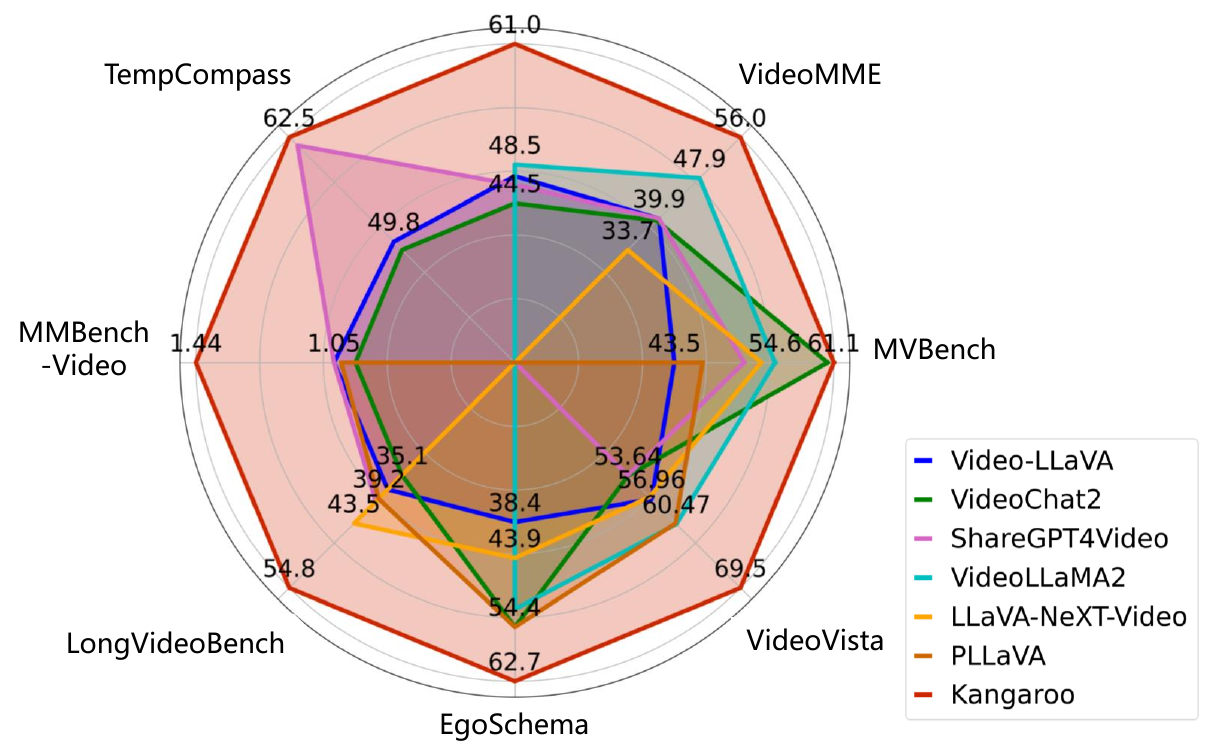}
  \caption{Kangaroo achieves SOTA performance across various comprehensive video understanding benchmarks, surpassing other 7B/8B models.}
  \label{radar}
\end{figure}

\begin{figure*}[htbp]
  \centering
  \includegraphics[trim=0 5.7cm 18cm 0, clip, width=0.75\linewidth]{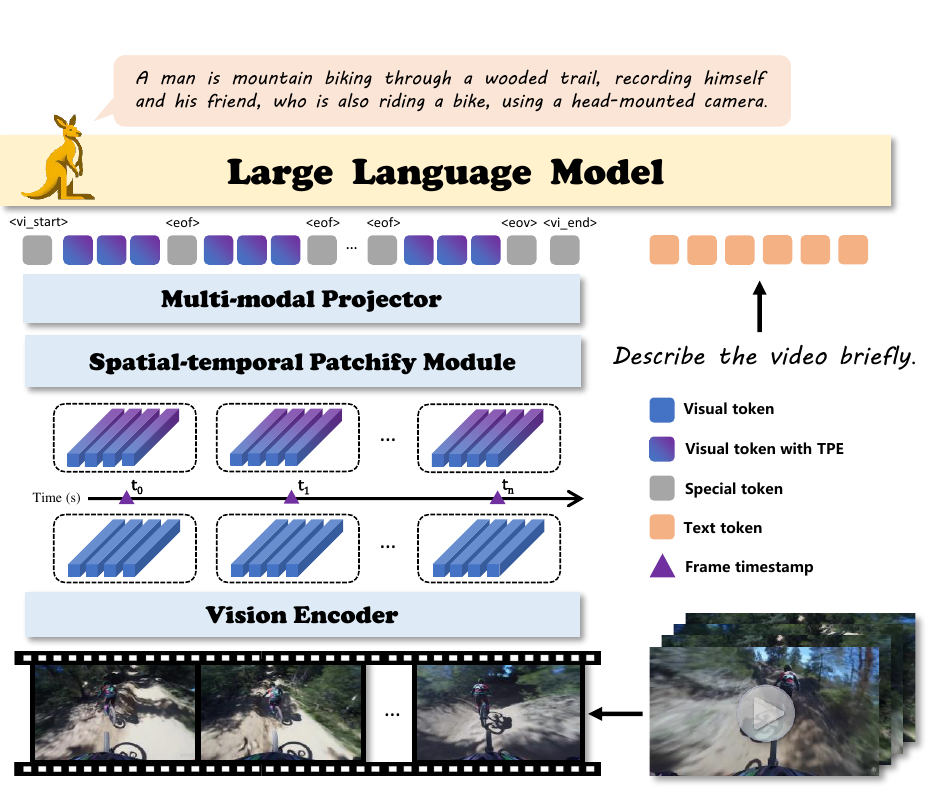}
  \caption{Architecture of Kangaroo. The model consists of a vision encoder, a spatial-temporal patchify module, a multi-modal projector and an LLM. Frames extracted from the input video are encoded by the vision encoder and augmented with temporal embedding. Special tokens, together with text tokens are embedded into the embedding space of LLM, and then fused with visual features before being fed to the LLM.}
  \label{model}
\end{figure*}

Leveraging the robust context processing and reasoning capability of Large Language Models (LLMs)~\cite{chatgpt, llama3, glm2024chatglm, cai2024internlm2, yang2023baichuan}, Large Multi-modal Models (LMMs)~\cite{gpt4v, liu2024visual, chen2024internvl, bai2023qwenvl, wang2023cogvlm} have made substantial progress in vision-language tasks. As a critical medium of visual information, video plays an indispensable role in interacting with real-world scenarios. Pioneer studies~\cite{lin2023videollava, maaz2023video, xu2024pllava, zhang2024llavanextvideo} begin to explore the extension of video inputs in LMMs. Despite demonstrating potential on certain benchmarks, existing methods still struggle with effectively processing long videos. We conduct an in-depth analyze of related research and identify two primary factors contributing to this issue. First, the scarcity of high-quality video datasets limits the efficacy of current approaches, leading to insufficient video-language alignment and suboptimal performance in video-related benchmarks. Second, numerous methods, as exemplified by studies such as~\cite{li2023videochat, li2024mvbench, jin2024chat, li2023llamavid}, incorporate aggressive resampling modules to exceedingly reduce the number of visual tokens. This aggressive compression results in a substantial loss of critical visual information, further exacerbating the problem.

In this paper, we propose \textbf{Kangaroo}, a powerful Video LMM aimed at ameliorating these issues. To overcome the obstacle posed by the limited availability of video data, a high-quality dataset is established for vision-language pre-training and instruction tuning. Utilizing this curated dataset, we devise a curriculum training strategy that progressively endows the LLM basement with the capacity to comprehend long videos. Moreover, recognizing the significance of visual information, we increase the resolution and number of input frames to improve the perception of global context and visual details in long videos. The evaluation results across diverse benchmarks for video understanding verify the superior performance of our model. Specifically, our key contributions in this paper are summarized as follows:

\begin{itemize}
    \item{\textit{\textbf{Data Curation System}:} We develop a data curation system encompassing data collection, filtering and caption generation. Initially, we collect a substantial amount of image and video data from various sources. To ensure high-quality video pre-training, low-quality videos are excluded prior to the implementation of the video caption generation procedure. Additionally, we establish a video instruction tuning dataset that spans multiple tasks, with task-specific prompts designed to enhance the model's instruction-following ability.}
    
    \vspace{3pt}
    \item{\textit{\textbf{Curriculum Training}:} We design a curriculum training strategy that is characterized by progressively increasing complexity and difficulty. The process initiates with image-text pre-training to establish fundamental vision-language correlations, followed by video-text pre-training to incorporate temporal perception. After pre-training, a high-quality subset is curated for refinement using high-resolution inputs across both spatial and temporal dimensions. Subsequently, we increase the number of input frames and conduct comprehensive fine-tuning across multiple instruction-following tasks. Finally, an additional phase is introduced, utilizing extensive input frames specifically optimized for longer videos.}

    \vspace{3pt}
    \item{\textit{\textbf{Long Context Input}:}
    To accommodate longer videos during instruction tuning, the maximum number of input frames is incrementally increased to 64 and 160, corresponding to context lengths of 10K and 22K tokens, respectively. With higher resolution and extensive input frames, our model is expected to effectively process extremely long videos.}
        
    \vspace{3pt}
    \item{\textit{\textbf{Outstanding Performance}:} We evaluate on various video understanding benchmarks. The results indicate that, with 8B parameters, Kangaroo achieves state-of-the-art performance on most benchmarks and reaches a competitive level on others. Notably, on benchmarks customized for long videos, Kangaroo outperforms certain larger models with over 10B parameters and proprietary models.}
    
\end{itemize}

\section{Related Work}

\subsection{Large Multi-modal Models}
The extraordinary performance of LMMs has greatly propelled the development of multi-modal models. Pioneer studies primarily concentrate on developing techniques for modality alignment. Kosmos-1~\cite{huang2024language} implements an end-to-end framework that directly merges visual inputs into LLM in a unified training process. Flamingo~\cite{alayrac2022flamingo, awadalla2023openflamingo} integrates visual features with linguistic features in each layer of LLM through a cross attention mechanism. BLIP-2~\cite{li2023blip} designs a Q-Former module to fuse visual and linguistic features via learnable queries. MiniGPT-4~\cite{zhu2023minigpt} and LLaVA~\cite{liu2024visual,liu2024improved} circumvent complex cross-modal fusion modules and by directly projecting visual features into the LLM embedding space through a lightweight multi-layer perceptron (MLP). 

Subsequent works aim to explore the application of LMMs in wider multi-modal tasks. Shikra~\cite{chen2023shikra}, Kosmos-2~\cite{peng2024grounding}, MiniGPT-v2~\cite{chen2023minigpt} and CogVLM~\cite{wang2023cogvlm} integrate visual grounding tasks, significantly enhancing the spatial perception ability of LMMs. Moreover, mPlug-DocOwl~\cite{ye2023mplug} and Kosmos-2.5~\cite{lv2023kosmos} construct specialized datasets to optimize document understanding. Recent works aim to unify diverse tasks into a general model, such as the LLaVA-NeXT~\cite{li2024llavanext}, InternLM-XComposer2 ~\cite{dong2024internlm2}, InternVL~\cite{chen2024internvl} and Qwen-VL~\cite{bai2023qwenvl}. These general LMMs employ additional optimization policies, high-quality datasets covering multiple tasks and intricate training strategies to advance performance across comprehensive vision-language tasks.

\subsection{Large Multi-modal Models for Video Understanding}
Advancements have been made to extend LMMs to video understanding tasks. To accommodate video inputs, video LMMs typically extract frames from the input video and rearrange the frame features encoded by the vision encoder to serve as the final video features. Some works~\cite{li2023videochat, li2024mvbench, zhang2023videollama} operate on the Q-Former module proposed by BLIP-2 to aggregate visual features and text features. On the other hand, methods including Video-LLaVA~\cite{lin2023videollava}, Valley~\cite{luo2023valley} and MiniGPT4-Video~\cite{zhu2023minigpt} adhere to the architecture of LLaVA, directly concatenating all frame features and projecting them through a simple fully connected layer. 

Certain research efforts investigate the use of spatial or temporal downsampling modules to compress visual tokens. For instance, Video-ChatGPT~\cite{maaz2023video} employs a pooling module to downsample visual tokens along both temporal and spatial dimensions. Chat-Uivi~\cite{jin2024chat} represents videos with a set of dymamic visual tokens to percieve high-level and low-level semantic details. Additionally, PLLaVA~\cite{xu2024pllava} introduces an adaptive pooling module for projected visual tokens to accommodate more video frames while minimizing the computational load. 
Recent LMMs~\cite{zhang2024internlm, li2024llavanext, chen2024internvl1.5} expand to support multi-image inputs. This enhancement enables the models to leverage extensive existing image-related datasets to elevate video comprehension.

\subsection{Long Video Understanding}
The methods mentioned above impose restrictions on the number of input frames, presenting challenges for understanding long videos. To handle hour-long videos, LLaMA-VID~\cite{li2023llamavid} adopts a context attention module to integrate multi-modal features and downsamples visual tokens into two tokens for each frame. MovieChat~\cite{song2024moviechat} designs a short-term and long-term memory module to consolidate local details and overall contents within the video. Similarly, TimeChat~\cite{ren2024timechat} introduces a timestamp-aware frame encoder to bind visual content with temporal information and uses a sliding Q-Former to integrate video tokens. Although these methods intend to enhance the perception of both local and global information in long videos, the excessive compression of visual tokens results in suboptimal results.

In contrast, some research adopts a stepwise strategy to incrementally adapt to long video inputs. LWM~\cite{liu2024world} gradually extends the context size from 4K to 1M tokens by optimizing the implementation of Blockwise RingAttention~\cite{liu2023blockwise} and other features for millions-length sequence training. LongVA~\cite{zhang2024long} attempts to leverage the long context generalization capability of LLMs for extremely long videos. The multi-modal model is trained based on an LLM initially generalized to millions of tokens. Training on long-context sequences ensures the entirety of visual information. However, the limited access to high-quality video data remains an obstacle to achieving ideal performance. 

Up to date, long video understanding remains a challenging task. Our proposed method aims to preserve the fidelity of the original visual information and elevate performance in long video understanding tasks by large-scale video-text datasets within a curriculum learning framework.
\section{Data Curation}

\begin{figure*}[htbp]
  \centering
  \includegraphics[trim=0 7cm 8.5cm 0, clip, width=0.9\linewidth]{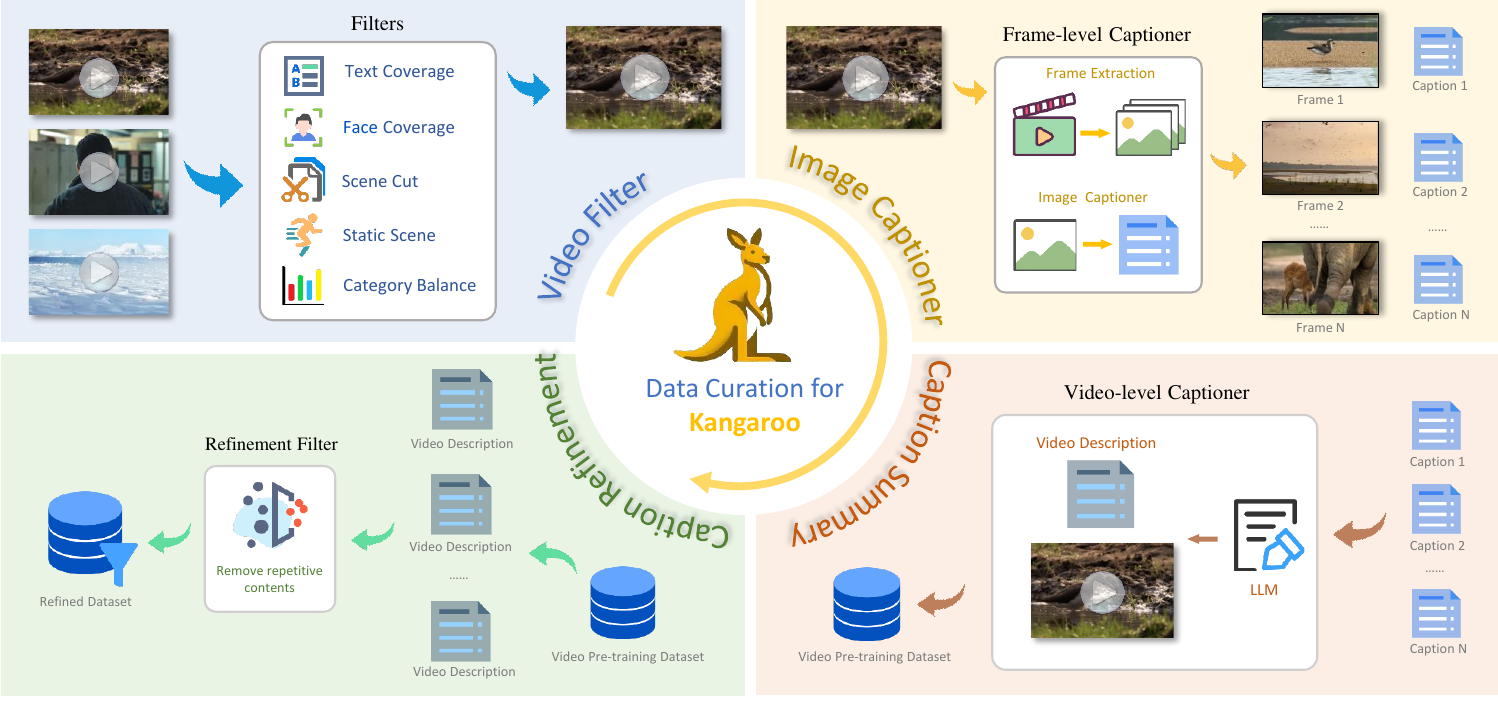}
  \caption{We develop a sophisticated data curation system tailored for video pre-training. To ensure high-quality input, we introduce a series of filters aimed at excluding low-quality videos. An off-the-shelf multi-modal model is subsequently employed to generate captions for frames extracted from each video. Following this, we utilize a well-trained LLM to summarize all frame-level captions into a coherent video description. For the pre-training refinement stage, we implement an additional strategy to discard video descriptions exhibiting repetitive content.}
  \label{curation}
\end{figure*}

In this section, we provide a detailed exposition of the data curation system. We start with the introduction of the image re-captioning procedure. Next, we elaborate on detailed steps of the video preprocessing and caption generation pipeline, setting the stage for robust video-text pre-training. In the last part, we demonstrate the composition of the multi-task instruction tuning dataset in detail. The schematic of our data curation process is depicted in Figure \ref{curation}.

\subsection{Image Re-Captioning}
Images form the foundation for bridging vision and language modalities. However, the captions accompanying images in existing public datasets are often noisy and unsuitable for pre-training. To address this issue, we implement a re-captioning procedure to create a high-quality dataset for image pre-training. Specifically, we select images from LAION-5B-en~\cite{schuhmann2022laion} for English data and Wukong~\cite{gu2022wukong} for Chinese data. The original captions associated with these images are discarded due to low quality. Instead, we employ an open-source multi-modal model to generate high-quality captions. Following a detailed evaluation of several off-the-shelf models considering factors including completeness, diversity, hallucination, image-text relevance and response speed, we ultimately opt for InternLM-XComposer2~\cite{dong2024internlm2} as the image captioner. The prompt is crafted to limit the generated English and Chinese captions within a preset length.

\subsection{Video Preprocessing}
To build a high-quality dataset for video pre-training, we gather videos from Webvid~\cite{Bain21}, Panda-70M~\cite{chen2024panda70m} with English captions and Youku-mplug~\cite{xu2023youku}, ChinaOpen~\cite{chen2023chinaopen} with Chinese captions. In addition, a number of internal videos are incorporated to enrich the diversity of video data.
However, subsequent examination of the large amount of video data reveals the low quality of certain videos. To mitigate the degradation of model performance caused by the inclusion of these videos, we propose several video filtering mechanisms, as outlined below:

\vspace{3pt}
\subsubsection{Text Coverage Filter} A common indicator of low-quality videos is the occlusion by large areas of irrelevant text elements, complicating the task of describing the video content accurately. Thus, we extract three frames from each video and employ PaddleOCR~\cite{du2022svtr} to detect text regions in these frames. Given the text bounding boxes $\{B_1, B_2, \ldots B_n\}$ in the i-th frame, the text coverage area of this frame $t_i$ is represented as the union area of all text regions, denoted as:
\setlength{\jot}{8pt}
\begin{align}
    t_i &= Area\left(\bigcup_{j=1}^{n} B_j\right)
\end{align}

The text coverage area of each video is defined as the maximum text coverage area among the three frames. Under this measurement criterion, we remove videos with a text coverage area surpassing the preset threshold.

\vspace{3pt}
\subsubsection{Face Coverage Filter} Another characteristic of inferior video quality is the the excessive proportion of facial content, as some videos feature bloggers speaking or interacting with the audience in front of the camera. Despite the richness of audio content, the visual information in these videos tends to be highly homogeneous and of low relevance. Similar to text coverage, we design a filter to screen these low-quality videos with large facial areas. We uniformly extract five frames from each video and employ YOLOv8~\cite{Jocher_Ultralytics_YOLO_2023} to detect facial regions. Videos with excessively large face coverage area in these frames are excluded.

\vspace{3pt}
\subsubsection{Static Scene Filter} Videos primarily containing static scenes with minimal temporal dynamics provide insufficient variability and context for robust video-text pre-training. To quantify the temporal variability, we apply Optical Flow~\cite{horn1981determining} algorithm to calculate motion vectors between consecutive frames of the video. To accelerate this procedure, the videos are first resized to a low resolution of 128$\times$128. Videos with optical flow magnitude below a certain threshold are classified as close-to-static and filtered out from the dataset. 

\vspace{3pt}
\subsubsection{Scene Cut} The original datasets include videos spanning a wide range of durations, from short videos of only 1-2 seconds to the longest videos exceeding 20 minutes. As a consequence, we impose a limit on the video duration by removing overly short videos and segmenting short clips from long videos via PySceneDetect~\cite{Castellano_PySceneDetect}.

\vspace{3pt}
\subsubsection{Category Balance} 
\begin{figure*}[htbp]
    \centering
    \begin{minipage}{0.58\linewidth}
        \centering
        \includegraphics[width=0.85\linewidth]{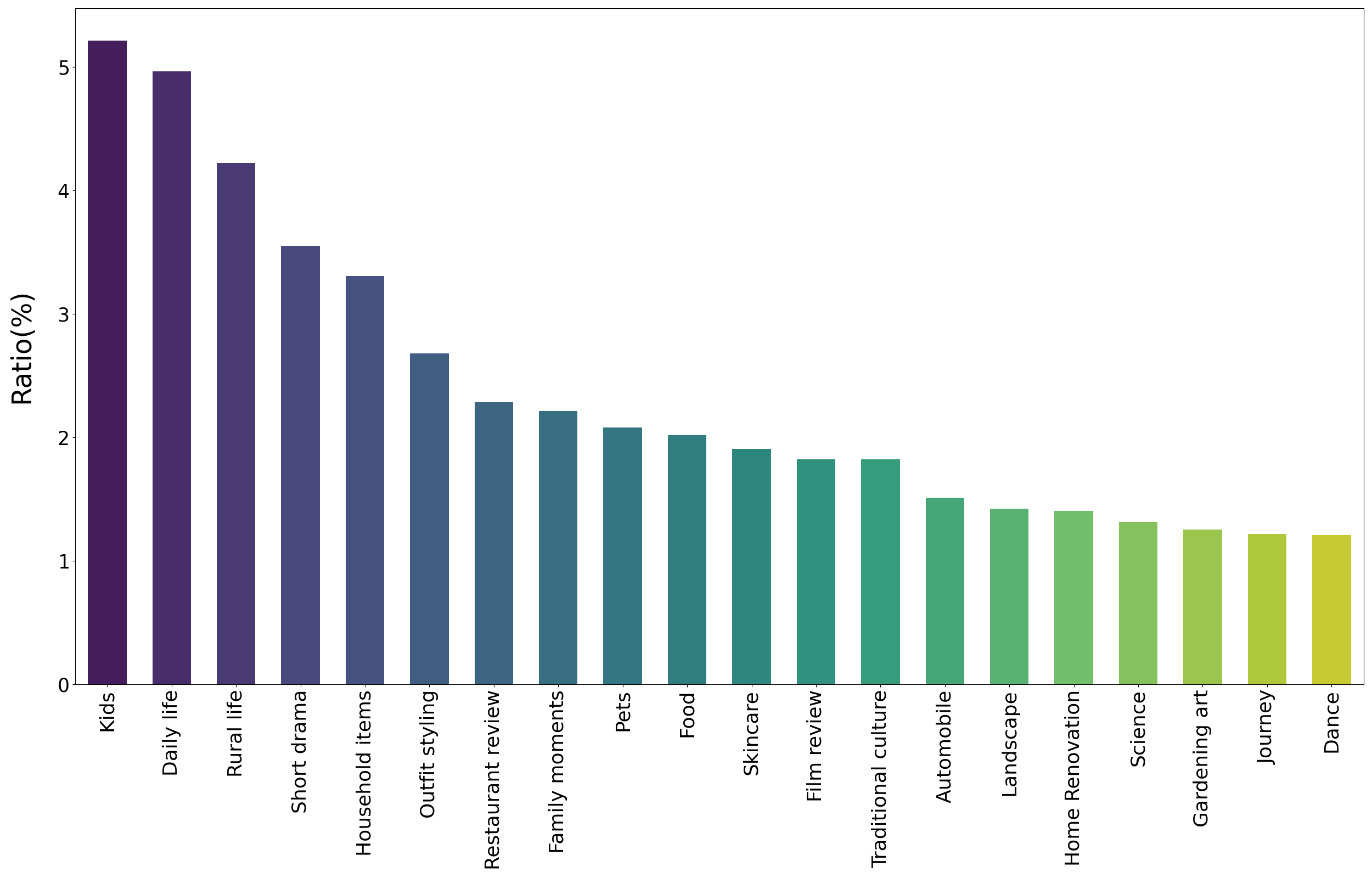}
        \caption{Distribution of Video Pre-training Dataset After Category Balance (Top 20) }
        \label{class balance}
    \end{minipage}
    \begin{minipage}{0.38\linewidth}
        \centering
        \includegraphics[trim=0 8.5cm 23.4cm 0, clip, width=0.75\linewidth]{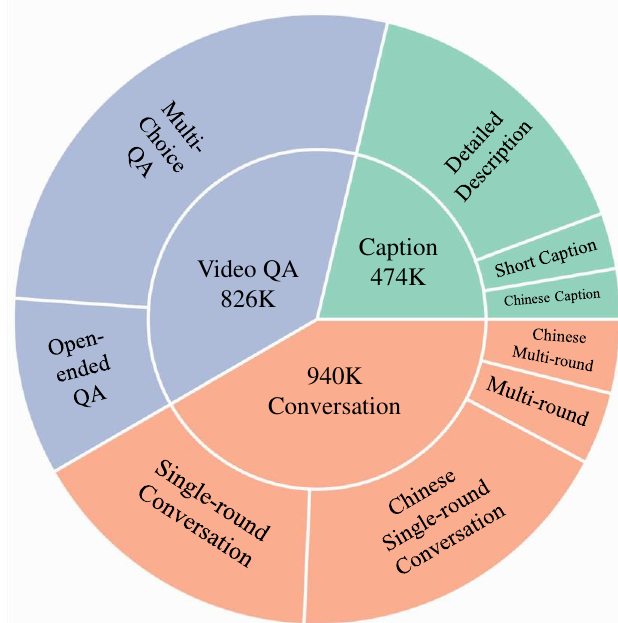}
        \vspace{16pt}
        \caption{Data Distribution of Instruction Tuning Dataset}
        \label{instruction}
    \end{minipage}
\end{figure*}

The highly imbalanced category distribution in a dataset poses a long-tail challenge and affects the model's generalization capability. Confronted with this problem, we employ an internal out-of-box video classifier to annotate all videos. By resampling the majority categories that exceed 1\% of the total dataset, we reduce their proportion to below 1\%. This approach ensures a relatively balanced category distribution in the final dataset as demonstrated in Figure~\ref{class balance}.

\subsection{Video Caption Generation}
Existing video LMMs are deficient in description accuracy, richness and diversity. Here we designed a comprehensive video captioning pipeline to address these shortcomings. Firstly, we evenly divide each video into five segments and randomly sample one frame from each segment. Next, a multi-modal model is employed to generate caption for the extracted frames individually. Following that, an off-the-shelf LLM is leveraged to synthesize these frame-level captions into a cohesive video-level caption. In practice, we compare the results of several open-source LMMs and ultimately take InternLM-XComposer2~\cite{dong2024internlm2} as the frame-level captioner. To avoid introducing bias by relying on a single model, we randomly select either InternLM2-Chat-20B~\cite{cai2024internlm2} or Baichuan2-13B-Chat~\cite{yang2023baichuan} to serve as the caption summarizer. Specialized prompts are designed to instruct the LLM to generate high-quality holistic descriptions. Upon completion of the video preprocessing and caption generation procedures, we establish a large-scale video pre-training dataset comprising over 60 million video-caption pairs.

\vspace{3pt}
\subsection{Video Caption Refinement}
Despite the specialized design of summary prompt, we discover that some generated video captions do not meet our expectations. In certain instances, the LLM tends to neglect the instruction from the summary prompt and simply concatenates individual frame captions without semantically restructuring. In addition, in some cases, the captions of adjacent frames from a video are highly correlated, resulting in repetitive sentences in the LLM outputs. These captions contain semantic redundancy and fail to present informative description of the entire video.

To mitigate the impact of repetitive contents, we curate a subset of 15M high-quality data from the 60M video-text dataset to refine video pre-training. We propose a metric to measure the similarity between sentences in a caption. For a video caption with $N$ sentences, we use jieba~\cite{jieba} for Chinese word segmentation and split English sentence into words. Let $s_i$ represent the word set of the \textit{i}-th sentence in the caption. For each pair $\{s_i, s_j\}$ of sentences, the similarity between these two sentences is quantified by the Intersection over Union (IoU) metric, defined as:
\begin{align}
S_{ij} = \frac{|s_i \cap s_j|}{|s_i \cup s_j|}
\end{align}

In this way, the sentence similarity of the entire video caption is denoted as the maximum similarity of all sentence pairs in the caption. To avoid semantic redundancy in the video captions, we remove data with sentence similarity exceeding the predefined threshold. The filtered video-caption data, along with dense video caption data from ShareGPTVideo~\cite{zhang2024direct} is used for pre-training refinement.

\begin{table*}[htbp]
    \centering
    \caption{Summary of instruction tuning dataset}
    \renewcommand{\arraystretch}{1.4}
    \begin{tabular}{cccc}
    \hline
        \textbf{Stage} & \textbf{Type} & \textbf{Size} & \textbf{Source} \\ \hline
        Image Pre-training & Image-text pairs & 300M & LAION, COYO and Wukong \\ \hline
        Video Pre-training & Video-text pairs & 60M & Webvid, Panda-70M, Youku-mlpug, ChinaOpen and internal videos \\ \hline
        Pre-training Refinement & Detailed description & 6.9M & ShareGPTVideo, refined caption on internal videos \\ \hline
        \multirow{8}{*}{Instruction Tuning} & Short caption & 89.7K & YouCook2, VaTEX, Charades, TextVR \\
        ~ & Detailed description & 35.5K & ShareGPTVideo, ShareGPT4Video, ShareGPT-4o \\
        ~ & Chinese detailed description & 50K & Generated detailed description on internal videos \\
        ~ & Multi-choice video QA & 616.1K & TV-QA, Cinepile, NExT-QA, TGIF-QA(MC), CLEVRER, SSthV2, Kinetics-710 \\
        ~ & Open-ended video QA & 209.9K & Ego-QA, NExT-OE, TGIF-QA(OE), Webvid-QA, MovieChat \\
        ~ & Single-round conversation & 655.7K & VideoChatGPT, ShareGPTVideo \\
        ~ & Multi-round conversation & 35.6K & VideoChat, VideoChat2, Valley (VaTEX subset) \\
        ~ & Chinese conversation & 547.7K & Generated conversation on ChinaOpen and internal videos \\ \hline
         Long Video Tuning & Long videos & 700K & Videos longer than 64s and short videos from instruction tuning dataset \\ \hline
    \end{tabular}
    \label{dataset summary}
\end{table*}

\subsection{Instruction Tuning Dataset}
Instruction-following data improves the model's generalization ability to various tasks~\cite{instructblip}. To enable the model's versatility to different tasks, we compile a large-scale video instruction tuning dataset comprising 2.24M samples from distinct public and internal sources. Detailed composition of the dataset is reported in Table \ref{dataset summary}. The curated instruction tuning dataset covers a wide range of video understanding tasks and can be divided into the following categories:

\vspace{3pt}
\subsubsection{Video Caption}
including short video caption and detailed video description data in both English and Chinese. Videos with short captions are collected from VaTEX~\cite{wang2019vatex}, YouCook2~\cite{zhou2018towards}, Charades~\cite{sigurdsson2016hollywood}, TextVR~\cite{wu2024large} and detailed description data from ShareGPTVideo, ShareGPT-4o~\cite{cui2024sharegpt4o}, ShareGPT4Video~\cite{chen2024sharegpt4video}. Each video from VaTEX and Charades corresponds to multiple short captions. Consequently, we leverage GPT-4~\cite{gpt4} with custom designed prompts to summarize all short captions into one overall caption. In addition, for Chinese understanding, a subset of Chinese detailed description data from the video pre-training dataset is incorporated in this stage. We formulate corresponding instructions for each type of caption to convert captions into QA pairs.

\vspace{3pt}
\subsubsection{Video QA}
including multi-choice VQA and open-ended VQA data collected from various sources. This subset encompasses fundamental VQA: NExT-QA~\cite{xiao2021next}, TGIF-QA~\cite{jang2017tgif}, WebvidVQA~\cite{yang2021webvid}, egocentric VQA: Ego-4d~\cite{grauman2022ego4d}, reasoning VQA: CLEVRER~\cite{yi2019clevrer}, action recognition: Something-Something-V2~\cite{goyal2017something, mahdisoltani2018effectiveness}, Kinetics-710~\cite{li2022uniformerv2} and VQA with long videos from TV-QA~\cite{lei2018tvqa}, Cinepile~\cite{rawal2024cinepile}, MovieChat. For WebvidVQA, Ego-4d and action classification subsets, we adopt the partition and instructions from MVBench~\cite{li2024mvbench}. We additionally design specialized templates for multi-choice VQA data to combine each question and the corresponding options. 

\vspace{3pt}
\subsubsection{Conversation}
including single/multi-round conversation in both Chinese and English. We include ShareGPTVideo, Video-ChatGPT and integrate single-round conversations from VideoChat, VideoChat2 and Valley into multi-round format. To improve the Chinese comprehension ability of our model, we select a subset from the video pre-training dataset and and utilize GPT-4 to generate human-like conversations based on the high-quality Chinese video description.

\vspace{3pt}
After collection and organization, we perform a statistical analysis of the resulting dataset and illustrate the distribution in Figure \ref{instruction}. Data with video length less than 5 seconds is excluded and all videos are resampled to 4 fps to accelerate data loading during the training process. .
\section{Methodology}
In this section, we present the overall architecture of our Kangaroo model and systematically elaborate on our curriculum training strategy.

\subsection{Architecture}
We utilize a widely adopted multi-modal architecture that bridges visual and linguistic features through a lightweight linear projector. As depicted in Figure \ref{model}, Kangaroo is composed of a vision encoder, a multi-modal projector, a spatial-temporal patchify module and an LLM. 

We initialize the vision encoder from EVA-CLIP-L~\cite{fang2023eva} and choose Llama-3-8B-Instruct~\cite{llama3} as our base LLM. For each input video $V \in \mathbb{R}^{T \times 3 \times H \times W}$, we start with uniformly sampling a series of \textit{n} frames $\{f_0, f_1, f_2, \ldots , f_n\}$. Inspired by position embedding proposed in~\cite{vaswani2017attention}, we design a temporal position embedding (TPE) to encode the temporal order of extracted frames in the video, implemented by sinusoidal position encoding:
\begin{align}
    TPE(t) &= \begin{pmatrix}
        sin(\frac{t}{\theta^{0/d}}) \\
        cos(\frac{t}{\theta^{1/d}}) \\
        \ldots  \\
        sin(\frac{t}{\theta^{(d-2)/d}}) \\
        cos(\frac{t}{\theta^{(d-1)/d}})
    \end{pmatrix} \\
    \hat{Z_f^t} &= Z_f^t + TPE(t)
\end{align}
where $t$ is the timestamp of each frame and $\hat{Z_f^t}$ denotes the visual feature augmented with temporal information. Here we use the actual float-type timestamp of a frame instead of its index to incorporate the video's meta information. 

Ultimately, we concatenate visual features of all frames along the time dimension with special tokens interleaved among the sequence to model the temporal inter-dependencies. The resulting visual sequence is then projected into LLM embedding space via the multi-modal projector as the final output of the visual branch, denoted as $Z_V \in \mathbb{R}^{(h \times w \times n) \times c}$:
\begin{align}
    Z_V = Projector(\hat{Z_f^0} \oplus \hat{Z_f^1} \oplus \ldots \oplus \hat{Z_f^n})
\end{align}
where $\oplus$ stands for the concatenate operation.
\vspace{3pt}

\begin{figure*}[htbp]
  \centering
  \includegraphics[trim=0 9.4cm 1.9cm 0, clip, width=0.9\linewidth]{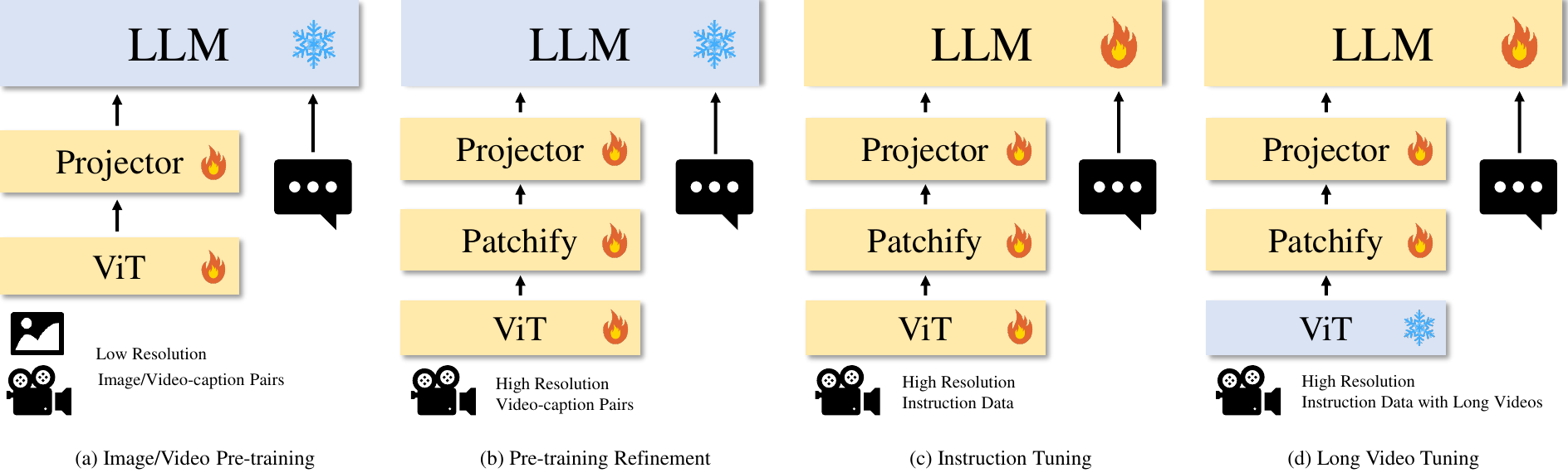}
  \caption{Training pipeline of Kangaroo. The training process is composed of five stages: (a) image pre-training stage and video pre-training stage, (b) pre-training refinement stage, (c) instruction tuning stage, (d) long video tuning stage. Trainable and frozen components in each stage are marked in the figure}
  \label{stages}
\end{figure*}

\begin{table*}[htbp]
    \centering
    \caption{Detailed training settings of Kangaroo}
    \renewcommand{\arraystretch}{1.3}
    \resizebox{0.95\textwidth}{!}{
        \begin{tabular}{cccccc}
        \hline
            \textbf{Configuration} &  \textbf{Image Pre-training} &  \textbf{Video Pre-training} &  \textbf{Pre-training Refinement} & \textbf{Instruction Tuning} & \textbf{Long Video Tuning}  \\ \hline
            ViT init & EVA02-CLIP-L-14 & Image Pre-training & Video Pre-training &  Pre-training Refinement & Instruction Tuning \\
            LLM init & Llama3-8B-Instruct & Llama3-8B-Instruct & Llama3-8B-Instruct & Llama3-8B-Instruct & Instruction Tuning \\
            Projector init & Random & Image Pre-training & Video Pre-training &  Pre-training Refinement & Instruction Tuning \\ 
            Patchify module init & - & - & Uniform &  Pre-training Refinement & Instruction Tuning \\ 
            Video/Image Resolution & 224$\times$224 & 224$\times$224 & 448$\times$448 & 448$\times$448 & 448$\times$448 \\
            ViT sequence length & 256 & 256 & 1024 & 1024 & 1024 \\
            LLM sequence length & 512 & 2560 & 2560 & 10K & 22K \\
            Maximum frame count & 1 & 8 & 16 & 64~(maximum) & 16064~(maximum) \\
            Warm-up steps & 1000 & 1000 & 0 & 0 & 0 \\
            Global batch size & 4096 & 128 & 128 & 128 & 128 \\ 
            Pipeline parallelism & 1 & 1 & 1 & 4 & 4 \\
            Sequence packing & \ding{55} & \ding{55} & \ding{55} & \ding{51} & \ding{51} \\
            Initial learning rate & 1e-4 & 3e-5 & 3e-5 & 1e-5 & 1e-5 \\
            Minimum learning rate & \multicolumn{5}{c}{1e-6} \\
            Gradient clip & \multicolumn{5}{c}{1.0} \\
            Numerical precision & \multicolumn{5}{c}{bfloat16} \\
            ViT Drop path rate & \multicolumn{5}{c}{0} \\ \hline
        \end{tabular}
        }
    \label{train}
\end{table*}

\subsection{Curriculum Training}
As shown in Figure \ref{stages}, we devise a curriculum training framework with progressively increasing task complexity and training difficulty to incrementally equips a text-based LLM with the capability to process long video inputs. 

\vspace{5pt}
\noindent\textbf{Stage \uppercase\expandafter{\romannumeral1}: Visual-language Pre-training. }
We start with image/video pre-training to connect fundamental language concepts and visual elements. First, we utilize the re-captioned image dataset for image pre-training. Each image is represented as a single-frame video and is randomly assigned a timestamp for temporal position embedding. We then extend to video pre-training based on the curated large-scale video-caption dataset. For each input video, we uniformly extract 8 frames and calculate the respective timestamps. During this stage, the visual encoder and multi-modal projector are trained via back-propagation, guided by the subsequent frozen LLM.

\vspace{5pt}
\noindent\textbf{Stage \uppercase\expandafter{\romannumeral2}: Pre-training Refinement. }
Due to the inconsistent quality of the initial video captions, we introduce an additional pre-training refinement stage. This stage aims to mitigate the effects of the fixed format and repetitive content present in the video pre-training dataset. To align with higher-quality data, the number and resolution of input frames are simultaneously increased to $16\times448\times448$. Furthermore, we implement a spatial-temporal patchify module using a 3D depthwise convolution operation to condense visual tokens, ensuring that the number of tokens fed into the LLM remains unchanged.

\vspace{5pt}
\noindent\textbf{Stage \uppercase\expandafter{\romannumeral3}: Instruction Tuning. }
After the preliminary alignment of visual and linguistic features, we fine-tune the entire model to enhance the instruction-following abilities across various tasks with the subsequent strategies:

\begin{itemize}
    \vspace{3pt}
    \item\noindent\textit{Visual Augmentation.}
    With the aim of preserving visual information, we maintain a high resolution for input frames and increase the maximum number of frames to 64. Under this setting, the maximum context length for the input of LLM is extended to 10K tokens in support of long-context video inputs.

    \vspace{3pt}
    \item\noindent\textit{Dynamic Frame Sampling.}
    We implement a a dynamic frame sampling strategy with a variable number of input frames. The number of input frames are initially set between 16 and 64, and then adjusted according to the video's duration. This approach allows for more frames to be captured from long videos to cover the overall content, while avoiding excessive frames sampled from short videos, which leads to information redundancy and reduces training efficiency. 

    \vspace{3pt}
    \item\noindent\textit{Sequence Packing.}
    Instead of padding all sequences to a standard length, we employ an advanced sequence packing strategy to process the fused multi-modal sequences of varying lengths. By aggregating multiple sequences of different lengths into one composite instance with appropriate attention masks, the sequence packing strategy reduces unnecessary computation of extra padding tokens within a batch.
\end{itemize}

\vspace{5pt}
\noindent\textbf{Stage \uppercase\expandafter{\romannumeral4}: Long Video Tuning. }
We introduce an additional stage to fine-tune the model with videos longer than 64 seconds selected from the instruction tuning dataset. The maximum number of input frames is further increased to 160 to accommodate extremely long videos. To maintain comprehension across different video lengths, a subset of short videos from the instruction tuning dataset is incorporated for balance. During this stage, we focus on generalizing the LLM to process longer context inputs, so we freeze the vision encoder and fine-tune other components.
\section{Experiments}

\begin{figure*}[htbp]
  \centering
  \includegraphics[trim=0 5.7cm 12.5cm 0, clip, width=0.9\linewidth]{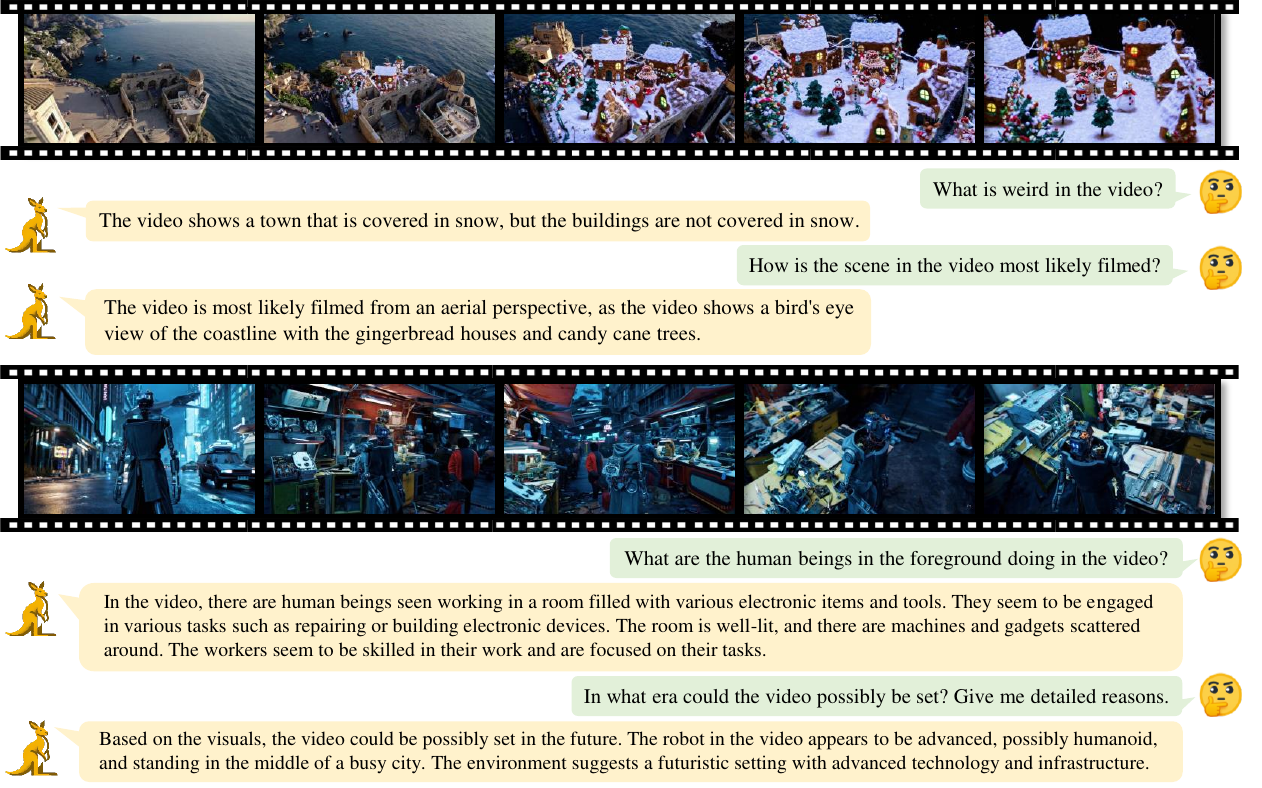}
  \caption{Qualitative results of reasoning capability. Kangaroo responds accurately and provides explicit reasons for the answer, such as inferring camera angles, recognizing actions and speculating about counterfactual scenarios.}
  \label{demo1}
\end{figure*}

\begin{figure*}[htbp]
  \centering
  \includegraphics[trim=0 5.2cm 12.9cm 0, clip, width=0.9\linewidth]{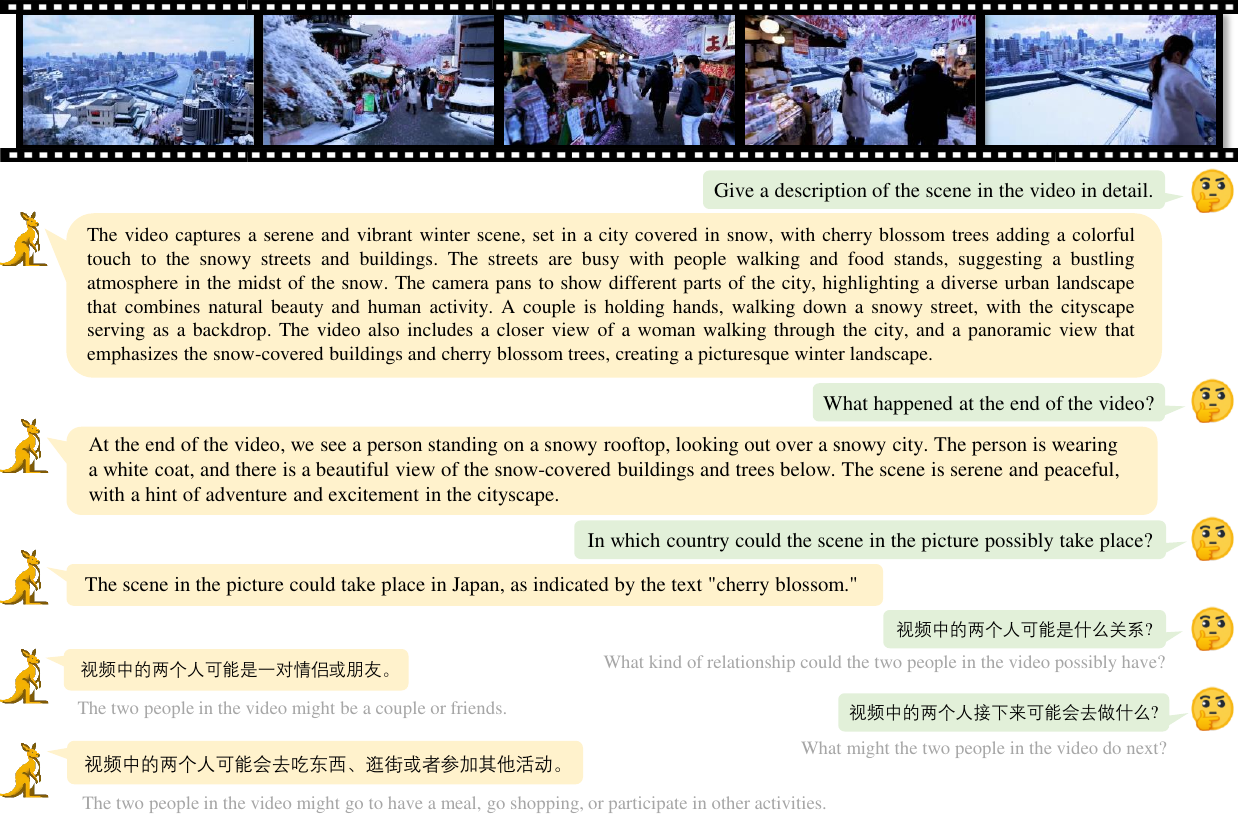}
  \caption{Qualitative results of multi-round and bilingual dialogues. Kangaroo exhibits a deep comprehension of contextual contents in multi-round dialogues and supports bilingual switching within the same conversation.}
  \label{demo2}
\end{figure*}

\begin{figure*}[htbp]
  \centering
  \includegraphics[trim=0 4.25cm 9cm 0, clip, width=0.95\linewidth]{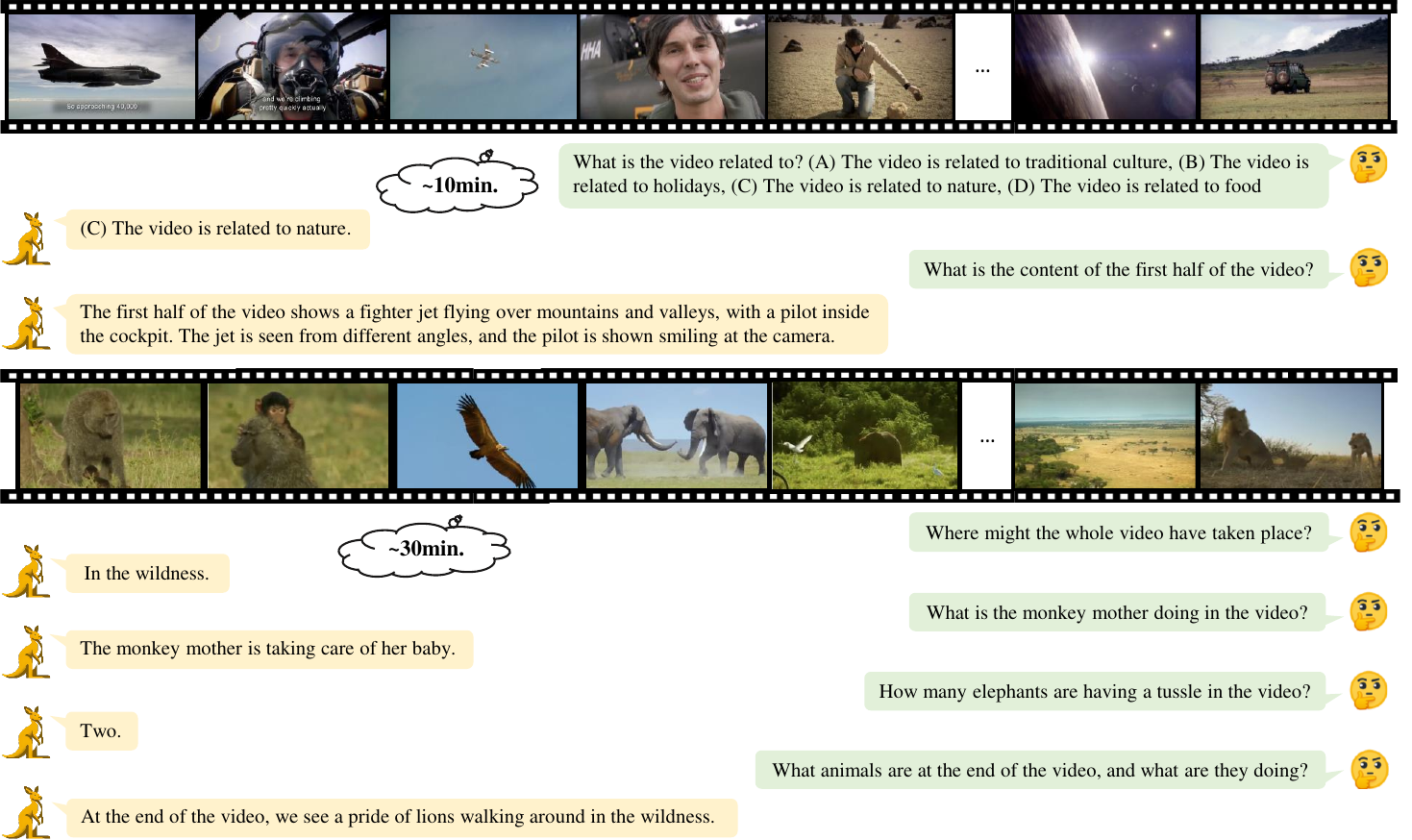}
  \caption{Qualitative results of long videos. Kangaroo effectively grasps the overall content in long videos (category, scene), precisely locates the relevant video segments and captures visual details (count, action) in specific segments from a long video.}
  \label{demo3}
\end{figure*}

\subsection{Implementation}
For all training stages, we use AdamW optimizer with the weight cay of 0.05 and cosine learning rate schedule. Layer-wise Learning Rate Decay strategy is applied for the vision encoder with the decay factor of 0.9. We only calculate cross-entropy loss for autoregressive text generation. Other detailed training settings are reported in Table \ref{train}.

\subsection{Evaluation}
We conduct standardized evaluation with greedy decoding and compare with other open-source and proprietary model. The best performance among the open-source models on each benchmark is highlighted in \textbf{bold} and the second-best results are \underline{underlined}. Items that Kangaroo outperforms proprietary models are denoted in \textcolor{gray}{gray}. 

To comprehensively explore the effectiveness of our model, we evaluate on a series of benchmarks covering a wide range of tasks, which can be categorized into three aspects: 

\vspace{3pt}
\subsubsection{General Video Understanding} We conduct evaluation on MVBench~\cite{li2024mvbench}, MMBench-Video~\cite{fang2024mmbench}, VideoVista~\cite{li2024videovista},  VideoMME~\cite{fu2024videomme} and SeedBench~\cite{li2024seed} and report the results in Table \ref{benchmarks}. MMBench-Video encompasses videos of varying durations with open-ended questions. VideoVista primarily assesses the model's reasoning ability across diverse content categories. Compared to other open-source LMMs with similar sizes, Kangaroo achieves \textit{state-of-the-art (SOTA)} performance on these two benchmarks, demonstrating our model's robust adaptability to different types of videos and strong reasoning capacity. Kangaroo approaches SOTA level and surpasses GPT-4V on MVBench, partly due to the relatively short video lengths in this benchmark. 

To illustrate the specific performance on different sub-tasks, we take VideoMME as an example. The detailed results in Table \ref{videomme} indicate that Kangaroo excels in all sub-tasks and outclasses some larger models with over 10B parameters. Additionally, we compare with other multi-image LMMs across video-related tasks from SeedBench. Given the short durations, only 8 frames of each video are provided in this benchmark, so we duplicate each set of 8 frames to 16 frames for evaluation. As demonstrated in Table \ref{seed}, with the same number of input frames, Kangaroo outranks most multi-image LMMs including Claude-3 and performs close to GPT-4V.

\vspace{3pt}
\subsubsection{Specialized Tasks} In addition to general video understanding benchmarks, we also test our model on the egocentric QA benchmark EgoSchema~\cite{mangalam2024egoschema} and the temporal reasoning benchmark TempCompass~\cite{liu2024tempcompass}. The results verify that as a general video understanding model, Kangaroo achieves SOTA performance among open-source models in these specialized tasks, and outperforms GPT-4V on EgoSchema.

\vspace{3pt}
\subsubsection{Long Video Understanding} We focus on benchmarks tailored for long videos, involving MLVU~\cite{zhou2024mlvu}, LVBench~\cite{wang2024lvbench} and LongVideoBench~\cite{wu2024longvideobench}. As reported in Table \ref{benchmarks}, Kangaroo considerably outperforms other 7B/8B models on these benchmarks. Notably, our model surpasses some larger models with over 20B parameters and proprietary models on certain metrics. This outstanding performance corresponds with our initial motivation, confirming the superiority of the curated high-quality dataset and curriculum training pipeline. 
\begin{table*}[htbp]
    \centering
    \caption{Evaluation results on comprehensive video understanding benchmarks}
    \renewcommand{\arraystretch}{1.4}
    \resizebox{0.9\textwidth}{!}{
        \begin{tabular}{cccccccccc}
        \hline
            \textbf{Model} & \textbf{Size} & \textbf{MVBench} & \textbf{MLVU} & \textbf{MMBench-Video} & \textbf{LVBench} & \textbf{EgoSchema} & \textbf{VideoVista} & \textbf{TempCompass} & \textbf{LongVideoBench} \\ \hline
            VideoChatGPT~\cite{maaz2023video} & 7B & 32.7 & 31.3 & 0.93 & - & - & 36.65 & 43.5 & - \\
            Video-LLaVA~\cite{lin2023videollava} & 7B & 43.5 & 47.3 & 1.05 & - & 38.4 & 56.96 & 49.8 & 37.6 \\
            VideoChat2-HD~\cite{li2024mvbench} & 7B & \underline{62.3} & 47.9 & - & - & \underline{55.8} & 61.58 & - & 41.2 \\ 
            LLaMA-VID~\cite{li2023llamavid} & 7B & 41.9 & 33.2 & - & 23.9 & 38.5 & 56.87 & 45.6 & - \\
            PLLaVA~\cite{xu2024pllava} & 7B & 46.6 & - & 1.03 & - & 54.4 & 60.36 & - & 39.2 \\ 
            VideoLLaMA2~\cite{cheng2024videollama} & 7B & 54.6 & 48.5 & - & - & 51.7 & 60.47 & - & - \\ 
            LLaVA-NeXT-Video~\cite{zhang2024llavanextvideo} & 7B & 53.1 & - & - & - & 43.9 & 56.66 & 53.7 & 43.5 \\ 
            LongVA~\cite{zhang2024long} & 7B & - & 56.3 & - & - & - & 67.36 & 56.9 & - \\
            TimeChat~\cite{ren2024timechat} & 7B & - & 30.9 & - & 22.3 & 33.0 & - & 50.6 & - \\ 
            ShareGPT4Video~\cite{chen2024sharegpt4video} & 8B & 51.2 & 46.4 & 1.05 & - & - & 53.58 & \underline{61.5} & 41.8 \\
            VILA-1.5~\cite{lin2024vila} & 8B & - & 56.7 & - & - & 50.4 & 64.18 & 58.8 & -\\
            Video-CCAM~\cite{tecent2024videoccam} & 9B & \textbf{64.60} & \underline{58.5} & - & - & - & \underline{69.00} & - & - \\ \hline
            \rowcolor{gray!4} InternVL-1.5~\cite{chen2024internvl1.5} & 26B & - & 50.4 & \underline{1.26} & - & - & - & - & - \\ 
            \rowcolor{gray!4} PLLaVA~\cite{xu2024pllava} & 34B & 58.1 & - & - & \underline{26.1} & - & - & - & \underline{53.5} \\
            \rowcolor{gray!4} VideoLLaMA2~\cite{cheng2024videollama} & 8×7B & 53.9 & 35.5 & - & - & 53.3 & - & - & - \\ \hline
            \rowcolor{gray!8} GPT-4V~\cite{gpt4v} & UNK & \textcolor{gray}{43.7} & \textcolor{gray}{49.2} & 1.53 & - & \textcolor{gray}{55.6} & - & - & 60.7 \\
            \rowcolor{gray!8} GPT-4o~\cite{gpt4o} & UNK & - & 64.6 & 1.63 & \textcolor{gray}{27.0} & 72.2 & 78.26 & - & 66.7 \\
            \rowcolor{gray!8} Gemini-1.5-Pro~\cite{team2023gemini} & UNK & - & - & \textcolor{gray}{1.30} &\textcolor{gray}{33.1} & 63.2 & 76.39 & 67.1 & 64.4 \\ \hline
            \textbf{Kangaroo (Ours)} & 8B & 61.1 & \textbf{61.0} & \textbf{1.44} & \textbf{39.4} & \textbf{62.7} & \textbf{69.50} & \textbf{62.5} &  \textbf{54.8}\\ \hline
        \end{tabular}
    }
    \label{benchmarks}
\end{table*}

\begin{table*}[!ht]
    \centering
    \caption{Comparison with other video LMMs on VideoMME benchmark}
    \renewcommand{\arraystretch}{1.3}
    \begin{tabular}{ccccccccccc}
    \hline
        \multirow{2}{*}{\textbf{Model}} & \multirow{2}{*}{\textbf{Size}} & \multirow{2}{*}{\textbf{Frames}} & \multicolumn{2}{c}{\textbf{Short}} & \multicolumn{2}{c}{\textbf{Medium}} & \multicolumn{2}{c}{\textbf{Long}} & \multicolumn{2}{c}{\textbf{Overall}} \\ \cline{4-11}
        ~ & ~ & ~ & w/o subs & w subs & w/o subs & w subs & w/o subs & w subs & w/o subs & w subs \\ \hline
        Video-LLaVA~\cite{lin2023videollava} & 7B & 8 & 45.3 & 46.1 & 38.0 & 40.7 & 36.2 & 38.1 & 39.9 & 41.6 \\
        Chat-Univi-v1.5~\cite{jin2024chat} & 7B & 64 & 45.7 & 51.2 & 40.3 & 44.6 & 35.8 & 41.8 & 40.6 & 45.9 \\
        VideoChat2~\cite{li2024mvbench} & 7B & 16 & 48.3 & 52.8 & 37.0 & 39.4 & 33.2 & 39.2 & 39.5 & 43.8 \\ 
        VideoLLaMA2~\cite{cheng2024videollama} & 7B & 16 & 56.0 & 59.4 & 45.4 & 47.6 & 42.1 & 43.8 & 47.9 & 50.3 \\
        LongVA~\cite{zhang2024long} & 7B & 128 & 61.1 & 61.6 & 50.4 & 53.6 & 46.2 & 47.6 & 52.6 & 54.3 \\
        ShareGPT4Video~\cite{chen2024sharegpt4video} & 8B & 16 & 48.3 & 53.6 & 36.3 & 39.3 & 35.0 & 37.9 & 39.9 & 43.6 \\
        SliME~\cite{zhang2024slime} & 8B & 8 & 53.3 & 55.4 & 42.7 & 44.4 & 39.8 & 41.7 & 45.3 & 47.2 \\ \hline
        \rowcolor{gray!4} Video-CCAM~\cite{tecent2024videoccam} & 14B & 96 & 62.1 & 63.9 & 52.8 & \textbf{55.9} & \textit{47.0} & 48.3 & 53.9 & \underline{56.1} \\
        \rowcolor{gray!4} InternVL-V1.5~\cite{chen2024internvl1.5} & 20B & 10 & 60.2 & 61.7 & 46.4 & 49.1 & 45.6 & 46.6 & 50.7 & 52.4 \\ 
        \rowcolor{gray!4} VITA~\cite{fu2024vita} & 8$\times$7B & 20 & \underline{64.2} & \underline{67.9} & \underline{53.3} & 55.3 & \textbf{47.6} & \textbf{49.6} & \underline{55.0} & \textbf{57.6} \\ \hline
        \textbf{Kangaroo (Ours)} & 8B & 64 & \textbf{66.1} & \textbf{68.0} & \textbf{55.3} & \underline{55.4} & 46.7 & \underline{49.3} & \textbf{56.0} & \textbf{57.6} \\ \hline
    \end{tabular}
    \label{videomme}
\end{table*}
\begin{table}[htbp]
    \centering
    \caption{Evaluation results on video-related tasks from SeedBench benchmark (*indicates results evaluated by self)}
    \renewcommand{\arraystretch}{1.4}
    \resizebox{0.48\textwidth}{!}{
        \begin{tabular}{ccccccc}
        \hline
            \textbf{Model} & \textbf{Size} & \textbf{GVU} & \textbf{AR} & \textbf{AP} & \textbf{PU} & \textbf{Avg.} \\ \hline
            InternLM-XC~\cite{zhang2023internlm} & 7B & 59.1 & 50.4 & 37.8 & 25.3 & 45.3 \\ 
            Qwen-VL-Plus~\cite{bai2023qwenvl} & 7B & 51.8 & 54.5 & 29.3 & \textbf{48.0} & 46.7 \\ 
            LLaVA-NeXT-IL*~\cite{li2024llavanext} & 7B & \textbf{62.2} & \underline{59.6} & \underline{53.0} & 32.4 & \underline{53.7} \\ 
            Mantis-Idefics2*~\cite{jiang2024mantis} & 8B & 60.7 & 57.0 & 46.0 & 37.5 & 51.9 \\ \hline
            \rowcolor{gray!4} SPHINX-v2~\cite{gao2024sphinx} & 13B & 52.4 & 41.2 & 33.9 & 26.1 & 40.0 \\ 
            \rowcolor{gray!4} VisionLLaMA~\cite{chu2024visionllama} & 13B & 58.3 & 49.2 & 43.8 & 42.3 & 49.4 \\ \hline
            \rowcolor{gray!8} Claude-3-Opus~\cite{claude3} & UNK & \textcolor{gray}{48.8} & \textcolor{gray}{37.1} & \textcolor{gray}{32.1} & \textcolor{gray}{41.3} & \textcolor{gray}{40.2} \\ 
            \rowcolor{gray!8} GPT-4V~\cite{gpt4v} & UNK & 64.5 & \textcolor{gray}{65.7} & \textcolor{gray}{51.7} & 63.4 & 61.7 \\ \hline
            \textbf{Kangaroo (Ours)} & 8B & \underline{61.4} & \textbf{69.7} & \textbf{54.6} & \underline{45.9} & \textbf{59.2} \\ \hline
        \end{tabular}
        }
    \label{seed}
\end{table}

\vspace{3pt}
\subsection{Qualitative Results}
We offer some qualitative illustrations in Figure \ref{demo1}. The questions are designed to assess the reasoning ability of our model. The results depict that Kangaroo is capable of responding accurately based on detailed visual elements in the video and providing explicit reasons, such as inferring camera angles, recognizing actions and speculating about counterfactual scenarios. Moreover, we present cases about multi-round and bilingual conversation in Figure \ref{demo2}. Kangaroo exhibits a deep comprehension of contextual contents in multi-round dialogues and supports transition between Chinese and English within the same conversation.

We also showcase some long video examples in Figure \ref{demo3}. The videos are selected from the long video benchmark MLVU, with durations of around 10 minutes and 30 minutes, respectively. Focusing the task for long videos, we specifically design questions to evaluate global and detailed comprehension of the video. The results demonstrate that Kangaroo is capable of grasping the overall content of long videos, such as identifying the category and scene of the video. Meanwhile, Kangaroo precisely locates the relevant video segments and responds based on specific segments from a long video.

\section{Conclusion}
In this paper, we present Kangaroo, a powerful Large Multi-modal Model that excels at long video understanding. We develop a data curation system and establish a large-scale video-based dataset. To improve the capacity for processing long videos, we adopt high-resolution inputs and extend the number of input frames to keep intact visual details. Extensive evaluation results demonstrate that Kangaroo achieves superior performance on diverse comprehensive video understanding benchmarks. Remarkably, on benchmarks specialized for long videos, Kangaroo outperforms larger models with over 10B parameters and proprietary models on certain metrics. 

Building upon these promising results, we envision future works focusing on the following aspects:
\begin{itemize}
    \item\noindent We aim to extend Kangaroo to additional modalities like audio, charts and interleaved image-video-text data. The integration of multi-modal information is expected to provide more robust and comprehensive understanding of real-world scenarios.

    \vspace{3pt}
    \item\noindent We intend to scale up the model size and equip with more powerful LLMs. Larger LLMs are expected to empower Kangaroo to capture complex patterns in data and improve multi-modal reasoning.
    
    \vspace{3pt} 
    \item\noindent We will investigate methods to generalize to more complex tasks. For challenging tasks including counting, visual grounding and temporal localization, specialized modules and high-quality data act as the vital role in pushing the boundaries of model's application domains.
\end{itemize}




\bibliography{references}
\bibliographystyle{IEEEtran}
\end{document}